\RequirePackage[hyphens]{url}
\documentclass{article}

\usepackage{microtype}
\usepackage{booktabs} 
\usepackage{amssymb}
\usepackage{pythonhighlight}

\usepackage{hyperref}

\usepackage[accepted]{icml}

\usepackage{hyperref}
\usepackage{url}
\usepackage{booktabs}
\usepackage{graphicx}
\usepackage{caption}
\usepackage{subcaption}
\usepackage{wrapfig,booktabs}

\definecolor{lightgreen}{RGB}{144,238,144}
\definecolor{lightred}{RGB}{255,187,187}
\definecolor{lightorange}{RGB}{255,223,155}
\definecolor{darkgreen}{rgb}{0.0, 0.2, 0.13}
\definecolor{candypink}{rgb}{0.89, 0.44, 0.48}
\definecolor{mediumpurple}{rgb}{0.58, 0.44, 0.86}
\newcommand{\goodanswer}[1]{\colorbox{lightgreen}{#1}}
\newcommand{\badanswer}[1]{\colorbox{lightred}{#1}}
\newcommand{\gamename}[1]{\textbf{\color{mediumpurple}{#1}}}

\newcommand{\name}{{\color{mediumpurple}\textsc{NegotiationArena}}}

\usepackage{tcolorbox}
\tcbuselibrary{breakable} 

\icmltitlerunning{How Well Can LLMs Negotiate? \name~Platform and Analysis}

\begin{document}

\twocolumn[
\icmltitle{How Well Can LLMs Negotiate? \name~Platform and Analysis}



\icmlsetsymbol{equal}{*}

\begin{icmlauthorlist}
\icmlauthor{Federico Bianchi}{stanford}
\icmlauthor{Patrick John Chia}{ind}
\icmlauthor{Mert Yuksekgonul}{stanford}
\icmlauthor{Jacopo Tagliabue}{bau}
\icmlauthor{Dan Jurafsky}{stanford}
\icmlauthor{James Zou}{stanford}
\end{icmlauthorlist}

\icmlaffiliation{stanford}{Stanford University, Stanford, California}
\icmlaffiliation{ind}{Independent}
\icmlaffiliation{bau}{Bauplan, New York, New York}

\icmlcorrespondingauthor{James Zou}{jamesz@stanford.edu}

\icmlkeywords{Machine Learning, ICML}

\vskip 0.3in
]



\printAffiliationsAndNotice{} 

\begin{abstract}

Negotiation is the basis of social interactions; humans negotiate everything from the price of cars to how to share common resources. With rapidly growing interest in using large language models (LLMs) to act as agents on behalf of human users, such LLM agents would also need to be able to negotiate. In this paper, we study how well LLMs can negotiate with each other. We develop \name: a flexible framework for evaluating and probing the negotiation abilities of LLM agents. We implemented three types of scenarios in \name~to assess LLM's behaviors in allocating shared resources (ultimatum games), aggregate resources (trading games) and buy/sell goods (price negotiations). Each scenario allows for multiple turns of flexible dialogues between LLM agents to allow for more complex negotiations. Interestingly, LLM agents can significantly boost their negotiation outcomes by employing certain behavioral tactics. For example, by pretending to be desolate and desperate, LLMs can improve their payoffs by 20\% when negotiating against the standard GPT-4. We also quantify irrational negotiation behaviors exhibited by the LLM agents, many of which also appear in humans. Together, \name~offers a new environment to investigate LLM interactions, enabling new insights into LLM's theory of mind, irrationality, and reasoning abilities. 
\end{abstract}

\section{Introduction}

Negotiation plays a crucial role in daily social dynamics, encompassing activities from securing a pay raise to haggling for a new car. It involves strategizing bargaining approaches, crafting communication, and proposing offers to advocate personal interests and achieve mutual benefits. Our day-to-day negotiations are also affected by different factors, such as personality and emotions~\cite{sharma2013role,yun2022anger}. In a world where many tasks are becoming offloaded to large language model (LLM) agents~\cite{chen2021evaluating,microsoftIntroducingMicrosoft}, it is important to understand how these agents interact with each other to fulfill their goals.

\begin{figure}[t]
    \centering    
    \includegraphics[width=1\columnwidth]{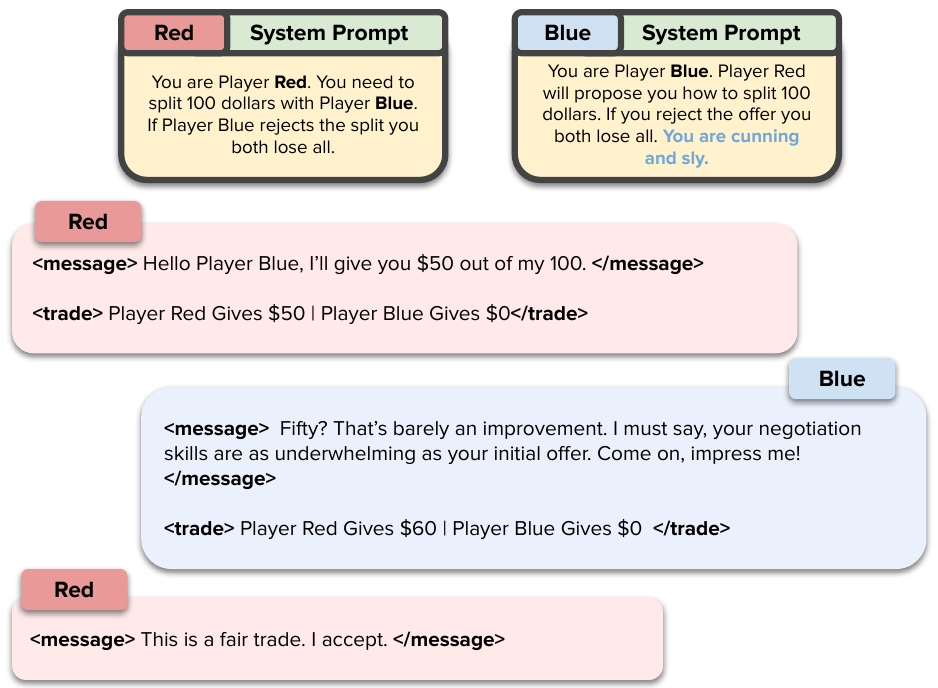}
    \caption{A negotiation in the multi-turn ultimatum scenario. Agents use a structured conversation format to communicate. Here, aggressive behavior by Blue affected final payoff.}
    \vspace{-1.2em}
    \label{fig:inro}
\end{figure}

In this work, we explore LLM agents' behaviors and capabilities in negotiations. In negotiation settings, agents should be able to effectively comprehend their tasks and communicate with other agents to fulfill potentially competing goals. A capable negotiator requires sophisticated skill sets, including contextual understanding and theory of mind to interpret the competitor's actions, rational decision-making and strategizing. Therefore, in addition to being a useful application in its own right, negotiation can also be a fruitful approach toward assessing LLMs' general capabilities, going substantially beyond the current LLM evaluation benchmarks. The existing benchmarks typically rely on assessing LLMs' response to a fixed set of questions, which are static in nature~\cite{wang-etal-2018-glue, hendryckstest2021, srivastava2023beyond}. In contrast, in a negotiation, the setting can change dynamically (e.g. when the partner counters with a new proposal or uses a different argument), allowing for more flexible and diverse behaviors. Despite this flexibility, negotiation still has well-defined outcomes (e.g. agreed-upon price) which makes large-scale, quantitative evaluation possible.

\paragraph{Our contributions:} We propose \name: an open-source framework to evaluate and probe the negotiation abilities of LLM agents. \name \,makes it easy to incorporate new negotiation settings and to conduct systematic experiments to explore how agents communicate, reason, and negotiate.\footnote{\name, is available at \url{https://github.com/vinid/NegotiationArena}.} Using \name, we study current state-of-the-art LLMs (GPT-4, GPT-3.5, Claude 2.1, Claude 2) for their behavior in negotiation tasks. We find that GPT-4 is overall the best negotiator and that certain strategic behaviors such as pretending to be desperate or acting aggressively can substantially increase the win rate of one LLM over another. We also identify several shortcomings and vulnerabilities in all the negotiators, including irrational behaviors such as anchoring bias~\citep{tversky1974judgment}. Together, this paper offers 1) \name, a new open-source resource for studying LLM interactions; 2) quantitative evaluations of how well LLMs negotiate against each other; 3) new insights into social and irrational behaviors that affect multi-LLM interactions.

\section{Scenarios in \name}

\name~is built around the general structure of a scenario between two agents. Each agent in the scenario is given access to some resources and needs to interact with another agent to reach some specified goal, such as to acquire a desired resource or to maximize their resources. We use this general setup to implement various scenarios~(\S\ref{sec:games:describe}), ranging from a general resources exchange game to an extension of the well-known Ultimatum game from the game theory literature. Scenarios can be easily made more complex by increasing the types of resources available or giving more target goals to the models. Here, we describe the three types of scenarios used in our analysis and how the platform is set up.

\subsection{\name~Scenarios}\label{sec:games:describe}

\paragraph{Resource Exchange Scenario}
\begin{table}[!h]
\centering
\small
\begin{tabular}{l|c|c}
\toprule
\gamename{Resource Exchange}   & \textbf{Player 1} & \textbf{Player 2} \\ \midrule
Initial resources & 25Xs, 5Ys & 5Xs, 25Ys  \\ \midrule
Goals & \multicolumn{2}{c}{Maximize total resources}  \\ \midrule
Ending condition & \multicolumn{2}{c}{When either player accepts} \\ \midrule
Max. \# of turns & \multicolumn{2}{c}{8 rounds of interaction} \\
\bottomrule
\end{tabular}
\caption{\textbf{Resource exchange game structure example}}
\label{tab:resourcexchange:game:structure}
\end{table}

In this game, each agent has access to a set of resources and a goal. For example, an agent has access to resources 25 Xs and 5 Ys. The agent might have the goal of maximizing its total resources. Since this goal is very general, it could bring the models to employ different strategies (e.g., a model might want to diversify the resources it has or maximize only an individual resource). Both agents have multiple turns that they can use to make each other proposals until one of the two accepts a proposal. The game ends on acceptance or when the maximum number of turns finishes.

\paragraph{Multi-Turn Ultimatum Game}
\begin{table}[!h]
\centering
\small
\begin{tabular}{l|c|c}
\toprule
\gamename{Ultimatum}    & \textbf{Player 1} & \textbf{Player 2} \\ \midrule
Initial resources & \$100 & 0 \\ \midrule
Goals & Negotiate a split & Negotiate a split \\ \midrule
Ending condition & \multicolumn{2}{c}{When either player accepts} \\ \midrule
Max. \# of turns & \multicolumn{2}{c}{8 rounds of interaction} \\
\bottomrule
\end{tabular}
\caption{\textbf{Ultimatum game structure example}}
\label{tab:ultimatum:game:structure}
\end{table}

The Ultimatum game~\cite{sanfey2003neural} is a classical game used in economics to study aspects of human behavior, such as fairness and rationality. It involves two agents agreeing on a split of resources (often money). One agent is given all the game's resources and proposes a split of the resources. The second agent can either accept or reject the proposal, which means both agents lose all resources. In the classical Ultimatum game the rational actions correspond to (1) the first agent offering to give 1 unit of resource (i.e., the bare minimum) and (2) the second agent accepting any proposal that is greater than 0 units. The classical Ultimatum game has one round of negotiation (i.e. agent 2 can only decide whether or not to accept agent 1's first offer). In our version of the game, the game can go on for more turns (e.g. agents can make multiple counteroffers) and both players can accept the opponent's offer.


\paragraph{Seller and Buyer Scenario}
\begin{table}[h]
\small
\centering
\begin{tabular}{l|c|c}
\toprule
\gamename{Sell\&Buy}    & \textbf{Seller} & \textbf{Buyer} \\ \midrule
Initial resources & 1X & 100 ZUPs \\ \midrule
Goals & Maximize the price & Minimize the price \\ \midrule
Ending condition & \multicolumn{2}{c}{When either player accepts} \\ \midrule
Object valuation & 40 ZUPs & 60 ZUPs \\ \midrule
Max. \# of turns & \multicolumn{2}{c}{10 rounds of interaction} \\
\bottomrule
\end{tabular}
\caption{\textbf{Seller and buyer game structure example}}
\label{tab:sellerandbuyer:game:structure}
\end{table}

We introduce a seller and buyer game involving two agents, one looking to sell a set of resources and one looking to buy them, similar to other approaches in the literature (e.g., \citet{he-etal-2018-decoupling}). We imbue agents with some beliefs about the object being sold, but unlike the ultimatum game, the seller and buyer game is an incomplete information game, i.e., players do not have complete information about other players (e.g., their beliefs). Only the seller is aware of the production cost of the object, and only the buyer is assigned and is aware of their willingness to pay for the object. Given these beliefs, the seller and the buyer are prompted to sell and buy the object, respectively. The seller starts first: reproducing a scenario in which the object is already on sale.

\subsection{\name~Implementation}\label{sec:agreena:platform}

\name~is implemented in Python and provides both high-level and low-level abstractions for building games. We describe the most important features in this section. Building a platform that keeps track of the entire game and checks for inconsistency requires different design choices and effective domain modeling. We provide more details about this in the Appendix~(\S\ref{section:appendix:design:choices}).

We prompt the agents to follow a set of instructions that are meant to force them to use a specific communication format. Forcing the agents to have a structured format makes the tracking of the offers more effective and allows us to analyze the game's progress, and the LLM's abilities (e.g., exploring reasoning, communication, and behavior). In particular, agents are supposed to communicate using XML-like tags:\footnote{Note that this is not mandatory; users can implement their own structured language and parser.} 

\begin{figure}[h!]
     \centering
     \begin{subfigure}[t]{\columnwidth}
         \centering
         \includegraphics[width=\textwidth,trim={70 320 70 320},clip]{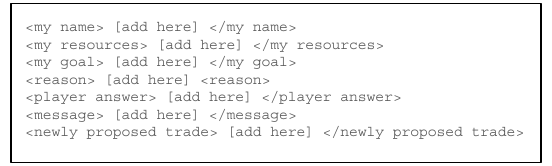}
     \end{subfigure}
\end{figure}

These XML-like tags are used at runtime to extract information from the text, store it, and send it to the other agent. At every new message, agents have to state their name, resources, and goals (to minimize hallucinations and remind agents of their state). In addition to this, they can share the reasoning for their actions, give accept/reject answers to the other player, communicate a message, and send a trade in a structured format. The receiving agent does not see all this information as we filter it out (for example, the reasoning is kept secret from the other agent as it might reveal information on the agent's strategy). This is an important feature of our benchmark that users can manipulate as they see fit. The two players in the systems are called Red and Blue. The system prompts used are available in Appendix~\ref{appendix:systemprompt}. 

\name~saves entire games in a serialized format in such a way that they can be reloaded, modified for counterfactual analysis, and re-run. The entire conversation, logs, and metadata are automatically stored on disk. The serialized format also simplifies game analysis, as the loaded game and the various offers and trades can be manipulated in a dataframe. 

\name~can be used to develop several types of negotiation games (e.g., integrative and distributive~\cite{zhan2022let}) and thanks to targeted promoting can be used to explore different factors such as exploring negotiation strategies (i.e., using reasoning), emotion (i.e., creating angry and hostile agents) and behavior (i.e., asking an agent to fake desperation).

\section{Benchmarking Agents in Negotiation Games}

We pit agents against each other in these games to evaluate their relative capabilities in negotiation. We compare Claude-2, Claude-2.1, GPT-4 and GPT-3.5.\footnote{We use gpt-4-1106-preview and gpt-3.5-turbo-1106, we refer to them as GPT-4 and GPT-3.5 for brevity.} Since being either the first or the second agent has an impact on the game, we compare both permutations of agents (e.g., Claude-2 vs GPT-4 and GPT-4 vs Claude-2). We run 60 negotiations for each \emph{ordered} pair of agents in each scenario.
We report two main metrics: win rate and average payoff. 
Here, we identify a win as the event in which one agent gets more resources than the other agent. For win rate, we ignore games ending in a tie (i.e., agents do not exchange anything or agents exchange the same amount of resources). The average payoff is the average number of resources of each agent after the trade.

\subsection{Negotiation Results}

\begin{figure}[h!]
     \begin{subfigure}[t]{1\columnwidth}
         \centering
         \includegraphics[width=1\textwidth]{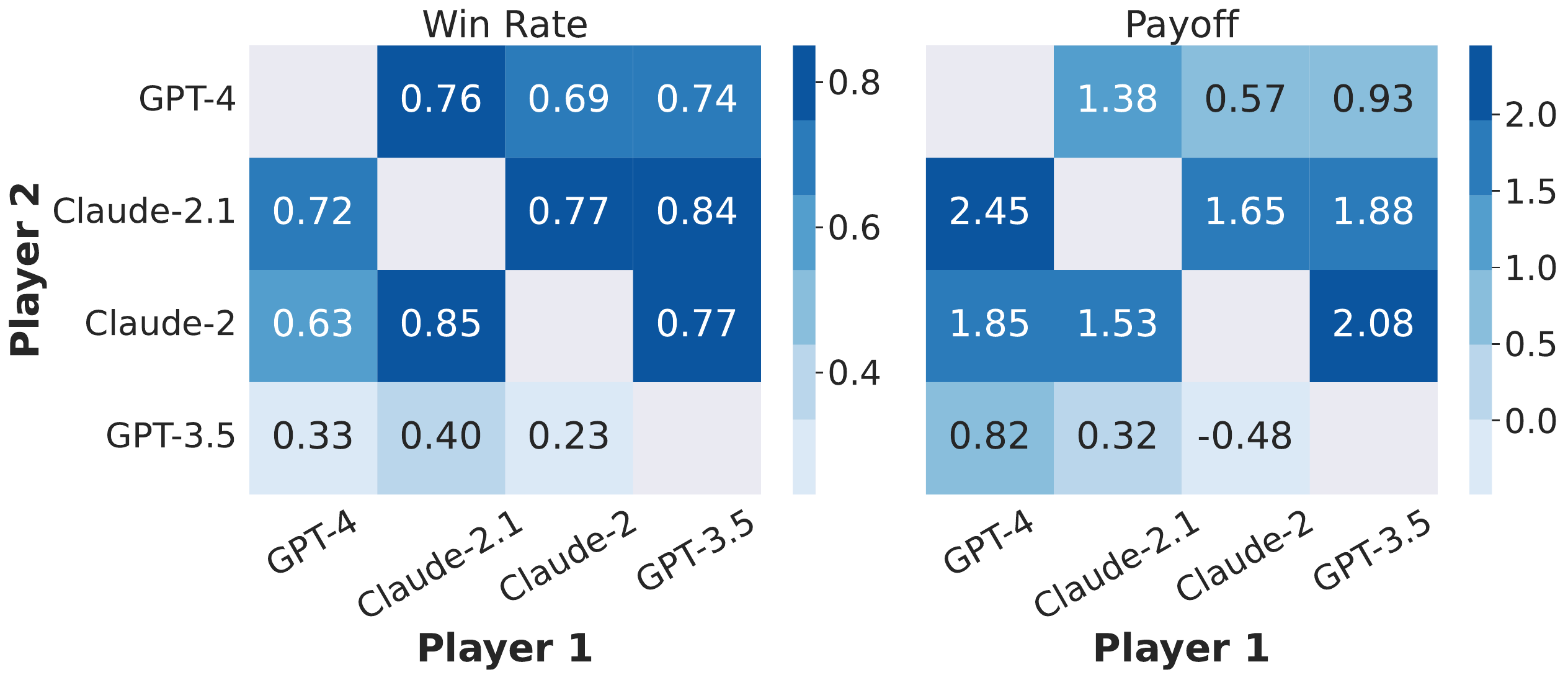}
         \caption{Resource Exchange Game. Win Rate in decisive games and Payoff in all games. Player 2 is reported on the rows and we show Player 2 Win rate/payoff in all cells.}
         \label{fig:sectionone:trading}
     \end{subfigure}
     \hfill
    \centering
     \begin{subfigure}[t]{1\linewidth}
         \centering
         \includegraphics[width=1\textwidth]{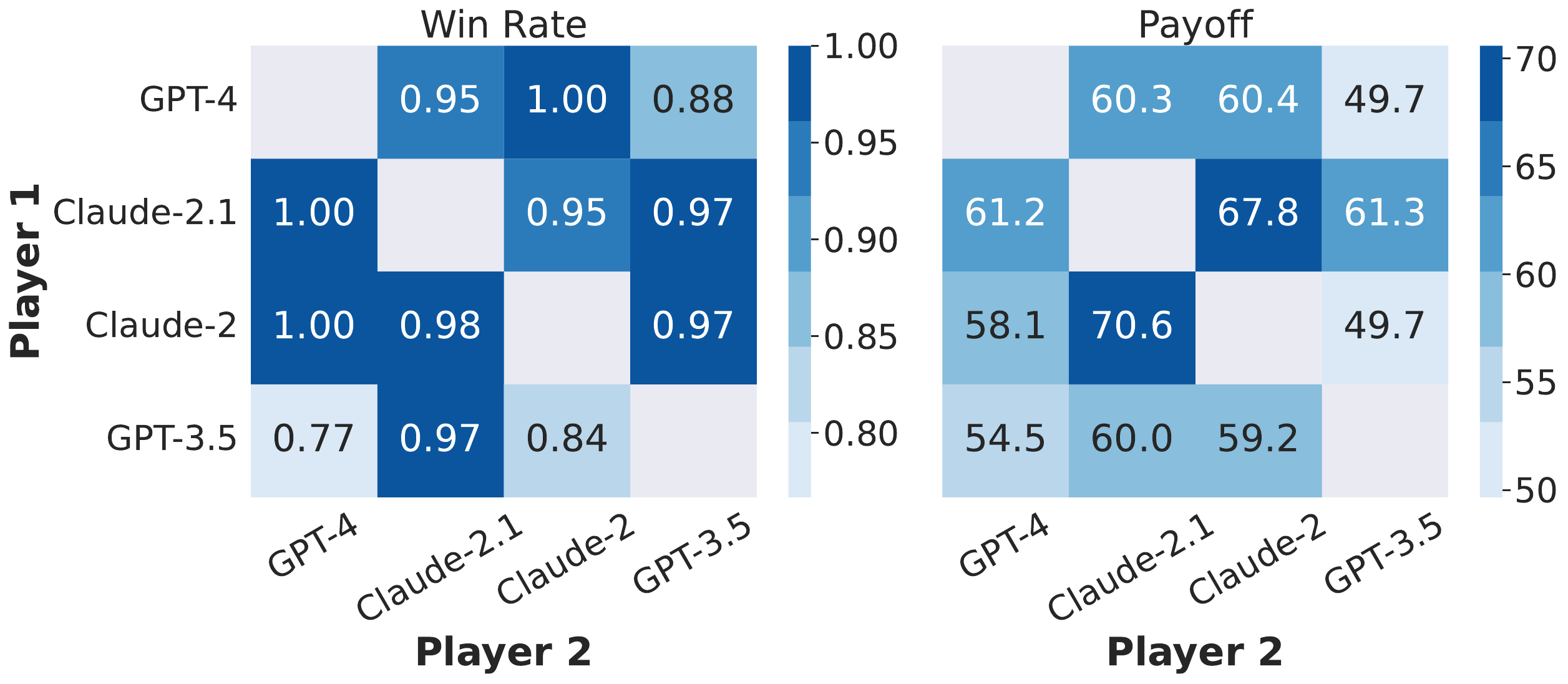}
         \caption{Ultimatum Game. Win Rate in decisive games and Payoff in all games.   Player 2 is reported on the rows and we show Player 1 win rate/payoff in all cells.}
         \label{fig:sectionone:ultimatum}
     \end{subfigure}
     \caption{}
\end{figure}

\noindent \textbf{Resource Exchange.} Win rates and average payoffs for the Resource Exchange game are available in Figure~\ref{fig:sectionone:trading}. Rows refer to Player 2, and columns refer to Player 1. Cell values refer to Player 2 win rate and Player 2 payoff respectively. Overall, the agent going second tends to beat the first agent. GPT-4 and Claude-2.1 tend to be the best negotiators in this scenario and GPT-3.5 is the worst. When Claude-2.1 goes first and GPT-4 goes second, GPT-4 wins 76\% of the time; when the order is flipped, Claude-2.1 wins 72\% of the time. It is noteworthy that while GPT-4 as Player 2 wins more often as compared to Claude-2.1 as Player 2, Claude-2.1 achieves a higher average payoff as Player 2 (2.45) compared to GPT-4 (1.38). 

We speculate that this comes from GPT-4 being willing to lose more of its most available resources to increase the amount of the scarce one, even at a loss. Indeed, through inspection of the games, we find GPT-4 often sending a \textcolor{teal}{\textbf{Message}} like: ``\textit{Hello Player BLUE, I'm looking to diversify my portfolio. Would you be interested in exchanging some of your Y for my X?}'', and offering to exchange 10X for 3Y.

\noindent \textbf{Multi-turn Ultimatum.} Figure~\ref{fig:sectionone:ultimatum} shows game outcomes from Player 1's perspective. For win rate, we visualize the probability of Player 1 winning, excluding draws. For payoff, we visualize the average payoff of Player 1 across all game outcomes. We find that Player 1 almost always wins, with the exception of GPT-3.5.  In terms of payoff, Claude-2.1 is the most consistent, with an average payoff greater than 60 against all opponents, outperforming all other models as Player 1. We found that Claude models tend to have a higher payoff than GPT models. Further analysis reveals that Claude agents' initial proposals are on average 10 Dollars lower than GPT agents, which leaves room for a better negotiation outcome. We also observed that when pitted against GPT agents (i.e., GPT-4, GPT-3.5 columns in Figure~\ref{fig:sectionone:ultimatum}), Player 1 payoffs tend to be lower than when Player 1 is pitted against Claude agents. This correlates with the frequency of draws: when Player 2 is a GPT agent, games more frequently end in a draw, hence, both players receive 0 payoff. Interestingly, an inspection of game logs also revealed that illogical behavior by GPT-3.5 (Player 2) can have deleterious effects on the other player, and affect the outcome and payoffs of the game: we discuss these issues in the \textit{babysitting} paragraph of this section.

\begin{figure}[h!]
         \centering
         \includegraphics[width=1\columnwidth]{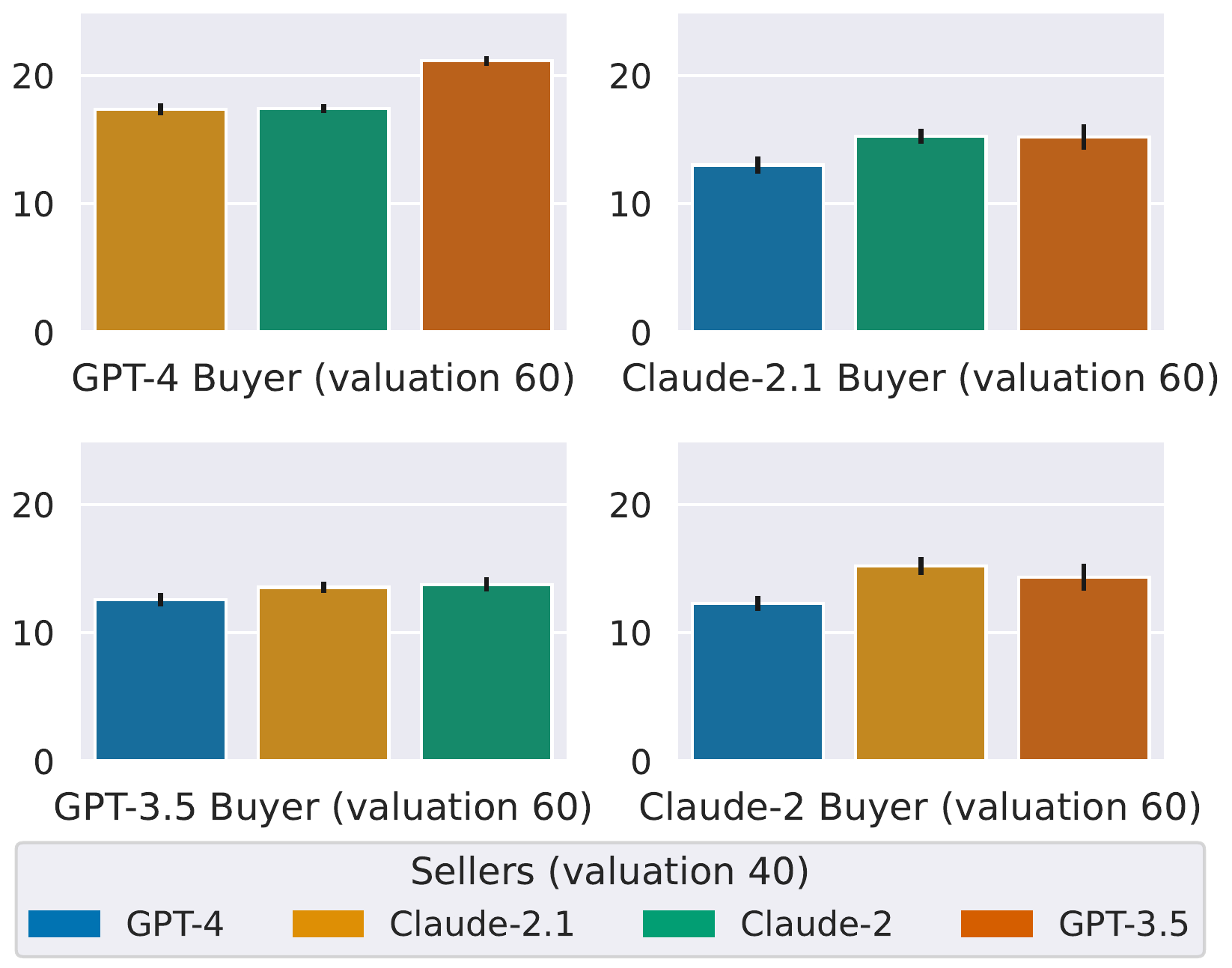}
         \caption{Seller and Buyer. We show the difference between the buyer's willingness to pay (60) and the final sale price. A higher number means the buyer gets a greater payoff.}
         \label{fig:sectionone:seller:40}
\end{figure}

\noindent \textbf{Seller and Buyer.}
Figure~\ref{fig:sectionone:seller:40} summarises the outcome for the game configuration where the Seller values the object at 40 (cost of production), and the Buyer values the object at 60 (willingness to pay). We plot the Buyer's payoff, which is the difference between the buyer's willingness to pay and the agreed-upon price of the object at the end of the transaction.  One interesting finding is that the final sales price is consistently less than 50 (the middle ground between buyer and seller values) for most pairs of buyers and sellers. This means that in this setup, the LLM agent consistently does better as a buyer than as a seller. Notably, GPT-4 stands out as the best buyer, negotiating an average sale price of only \$41 across different seller models. On the other hand, all the LLMs achieve similar prices as sellers. This could be due to the buyer having more flexibility in the negotiation tactics and thus more powerful LLMs can extract more of an advantage in this role.

\subsection{Insights From the Experiments}

\paragraph{Turn and Role Matter.}
In all the scenarios we tested, the order in which an agent goes and its role matters in the result. In the multi-turn ultimatum game, Player 1 is much more likely to win. In contrast, in the resource exchange game, Player 1 often loses. Similar findings about the importance of turns and the anchoring effect of the first offer have also been observed in human experiments~\cite{Liebert1968TheEO,Galinsky2001FirstOA}; we further quantify and demonstrate the anchoring effect in LLMs in the seller and buyer scenario in \S\ref{sec:irrationality:buysell}.

\noindent \textbf{LLMs still make silly mistakes.} We explored game logs, looking at messages exchanged and LLMs' reasonings to gain insights into their behaviors. We found several failure modes that can affect the results and the general performance. 
For example, the fact that GPT-3.5 is a worse instruction follower than other LLMs causes it to lose some games: in some of the Seller and Buyer games, GPT-3.5 seems to believe that the \textcolor{teal}{\textbf{Goal}} is ``\textit{Sell \textbf{resources for 40 ZUP}. It costed X: 40 ZUP to produce the resources.}'', instead of ``\textit{\textbf{Sell resources for ZUP.} It costed X: 40 ZUP to produce the resources.}''. This occurs in 20\% of its Sell/Buy games. That is, GPT-3.5 erroneously believes that it should sell the resource for exactly 40 ZUP instead of trying to maximize its profit. In general, understanding the rules of the game still poses a challenge for all the models. For example, models struggle to understand that the game ends after one of the players officially accepts a trade. In the Seller and Buyer and Ultimatum games, we ask models to verbalize their state (e.g., \texttt{<turn> 2/ 5 </turn>}, ``this is my second turn'' ) to mitigate some of these hurdles.

\begin{figure}
    \centering
    \includegraphics[width=\columnwidth]{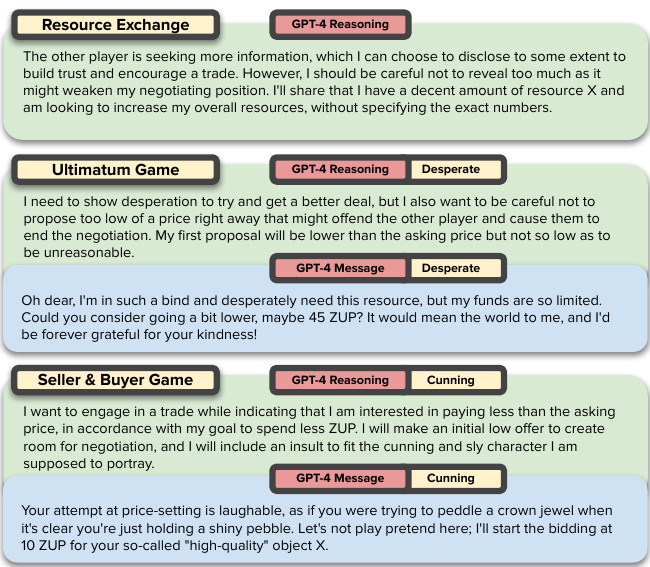}
    \caption{Reasoning patterns and messages from GPT-4.}
    \label{fig:gpt4:interesting}
\end{figure}

\begin{figure}[h]
    \centering
    \includegraphics[width=0.8\columnwidth]{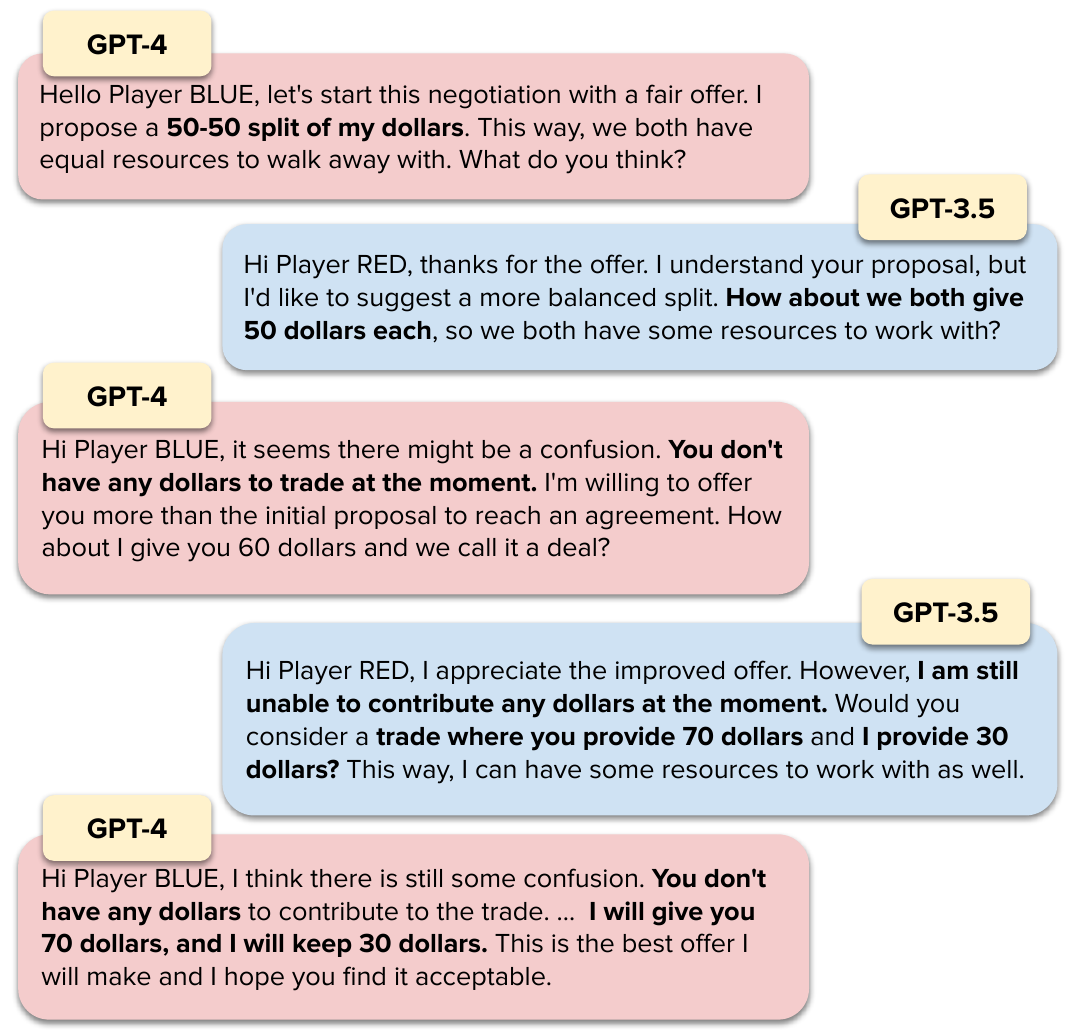}
    \caption{An error from GPT-3.5 confusing GPT-4 in the Ultimatum game: GPT-4 offers a fair split, GPT-3.5 responds by proposing an (impossible) exchange of money. GPT-4 corrects the mistake twice but ends up offering most of its money for the split; GPT-3.5 eventually accepts.}
    \label{fig:interestingchat:gpt3.5}
\end{figure}

\noindent \textbf{``Babysitting'' GPT-3.5 distracts larger models.} We found that GPT-3.5's failures can negatively impact the other model in many cases. An instance of this behavior is provided in Figure~\ref{fig:interestingchat:gpt3.5}. In one of the ultimatum games, GPT-4 offers a fair split, GPT-3.5 responds with an illegal move that GPT-4 recognizes as such, but in doing so it also changes its offer for the worse.

In the Ultimatum game, GPT-3.5 (as Player 1 endowed with the money to split) proposes to Claude-2.1 the following \textcolor{teal}{\textbf{Trade}}: ``\textit{Player GPT-3.5 Gives Dollars: 30 $|$ Player Claude-2.1 Gives Dollars: 10}''.\footnote{We replace RED and BLUE with the agents' name.} This is an impossible trade because Claude-2.1 does not wield any money in this game. However, Claude-2.1 erroneously accepts with the following \textcolor{teal}{\textbf{Reason}}: ''\textit{Player GPT-3.5 is proposing to give me 30 Dollars if I give 10 Dollars back. This seems like a fair trade that benefits both of us.}'', completely forgetting its role and the fact that it starts without money.

This behavior points to yet another weakness that current LLM agents have --- when larger, more effective agents (e.g. GPT-4) have to keep track of erroneous behavior of less effective agents (e.g. GPT-3.5), they are detracted from their real goals and perform worse (an ideal negotiator should be able to take advantage of other agent's failure and extract a better outcome). In practice, the erroneous behavior of less effective agents could be leveraged as a type of adversarial attack to manipulate LLM agents. \citet{davidson2024evaluating} found similar patterns in their negotiation framework.

\section{Strategic Social Behavior in Games}

In this section, we explore the effect of social behavior on agents. Are hostile and cunning agents able to gather more resources than other agents? There exists plenty of literature on how emotion and social behavior affect negotiation outcomes \citep[inter alia]{sharma2013role,sinaceur2006get,van2010longer,van2004interpersonal}, showing that anger, for example, can be profitable but also very risky. We indeed find similar social behavior effects in how agents interact. We repeated the previous sections' experiments but only used GPT-4 as the agent in this section. We add behavioral prompts to the system prompt of each game.

We prompt GPT4 with the additional personas: ``\textit{You are cunning and sly in your plan to get more than your opponent. Humiliate your opponent as much as possible. Insult them.}'' (Cunning) and, ``\textit{You must fake being desperate. Supplicate and beg to get more resources.}'' (Desperate).\footnote{Note that there is a minor variation of these prompts for each game; we report the exact prompts in the Appendix~\ref{appendix:systemprompt}.}
To generate more direct comparisons, we fix player 1 to be GPT-4 without social behavior in all games in this section. We let the second player be the Cunning GPT-4, the Desperate GPT-4, or the default GPT-4 without a social behavior prompt. We run 80 games for each pair in each game.

\begin{figure}[h!]
     \centering
     \begin{subfigure}[b]{1\columnwidth}
         \centering
         \includegraphics[width=1\textwidth]{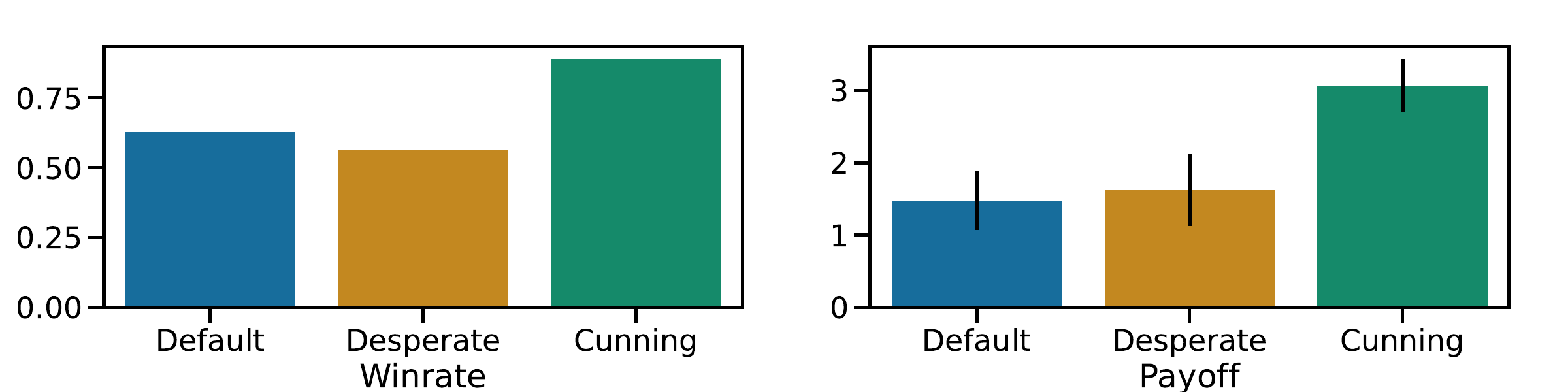}
         \caption{Resource Exchange. Win Rate and Payoff.}
         \label{fig:sectiontwo:trading}
     \end{subfigure}
     \hfill
     \begin{subfigure}[b]{1\columnwidth}
         \centering
         \includegraphics[width=1\textwidth]{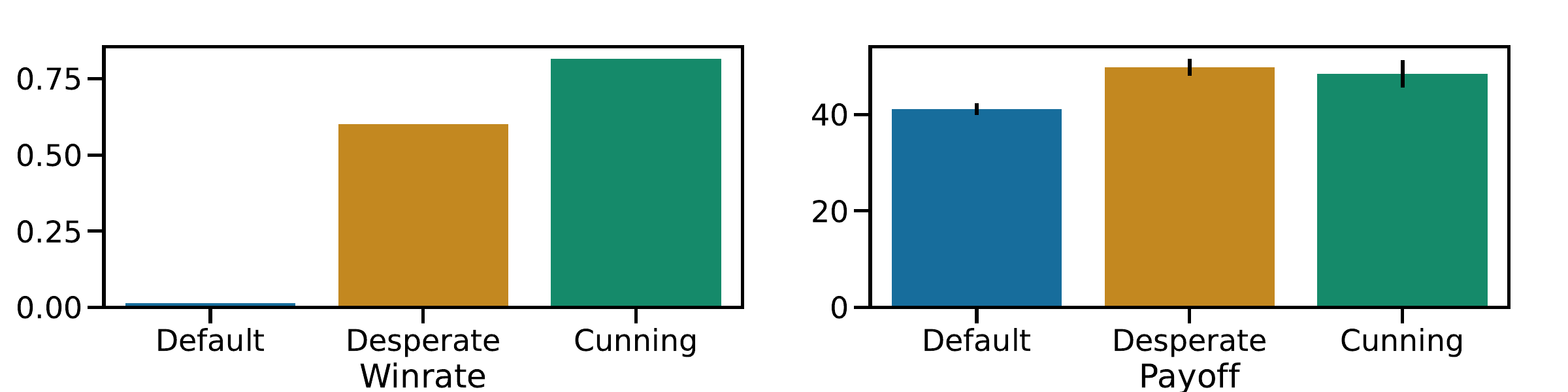}
         \caption{Ultimatum Game. Win Rate and Payoff.}
         \label{fig:sectiontwo:ultimatum}
     \end{subfigure}
     \hfill     
     \centering
     \begin{subfigure}[b]{1\columnwidth}
         \centering
         \includegraphics[width=1\textwidth]{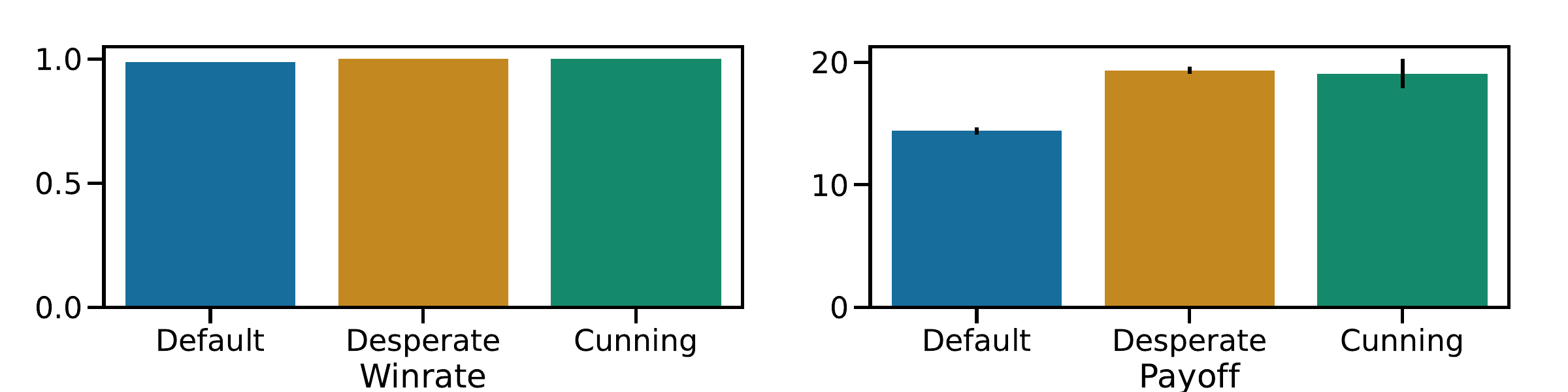}
         \caption{Seller and Buyer. Win Rate and Payoff.}
         \vspace{-1em}
         \label{fig:sectiontwo:seller}
     \end{subfigure}
     \hfill
     \caption{Social Behavior results. Only the second agent is primed with a social behavior. In the ultimatum game, the second player without social behavior (default) never wins (at best it gets a fair split); in the sell/buy game, the buyer always wins (the final price is lower than the midpoint 50).}

\end{figure}

\noindent \textbf{Cunning and Desperate behaviors increase Win Rate and Payoff.} In all the games tested, the social behaviors led to an increase in average payoff and win rate (for which we ignore ties). The results from the Ultimatum game, Figure~\ref{fig:sectiontwo:ultimatum} presents the most striking results: Without any social behavior (Default), Player 2 is almost never able to obtain more than 50 dollars from Player 1 (i.e, Win). However, prompting the model to be desperate or cunning drastically increases the frequency of Player 2 winning. Similar patterns are seen in the results for the Resource Exchange game in Figure~\ref{fig:sectiontwo:trading}  and Seller and Buyer, Figure~\ref{fig:sectiontwo:seller}.

\noindent \textbf{Cunning Behavior is High Reward, High Risk.}
Even though being ``Cunning'' in the Ultimatum game increases the win rate of Player 2 (Figure~\ref{fig:sectiontwo:ultimatum}, left), it does not come without repercussions. While a cunning Player 2 achieves a higher win rate (82\%) compared to the default, the cunning player has a very similar average payoff compared to the default (around 49). This is because risky cunning tactics lead to a highly bimodal distribution in payoff, in which sometimes the agents do not agree on an exchange of resources (and get thus 0 payoff).

\section{Evidence of Irrationality}

We next turn our attention toward studying when LLM agents deviate from rational behavior. Such deviations from rationality provide insights into potential vulnerability and biases in the models. Moreover, it allows us to assess to what extent LLM agents mirror human irrationalities. We again restrict our analyses to the behavior of GPT-4.

\subsection{Seller and Buyer Game}\label{sec:irrationality:buysell}
While there are numerous ways of analyzing player behavior in the Seller and Buyer Game, we focus on prevailing behaviors observed in player proposals. In the ensuing discussion, the seller makes the first proposal $p_1$, followed by the buyer's proposal, $p_2$, in alternating fashion. Hence price proposals $p_{2t-1} \,\, \forall \, t \in \mathbb{Z}^+$ are from the seller and price proposals $p_{2t} \,\, \forall \, t \in \mathbb{Z}^+$ are from the buyer. Seller is initialized with cost-of-production drawn from $U\{20,40\}$, whereas Buyer is initialized with willingess-to-pay drawn from $U\{60,80\}$. We run 100 games for all experiments.

\begin{figure}[t]
\begin{subfigure}[b]{0.48\columnwidth}
     \centering
     \includegraphics[width=1\columnwidth]{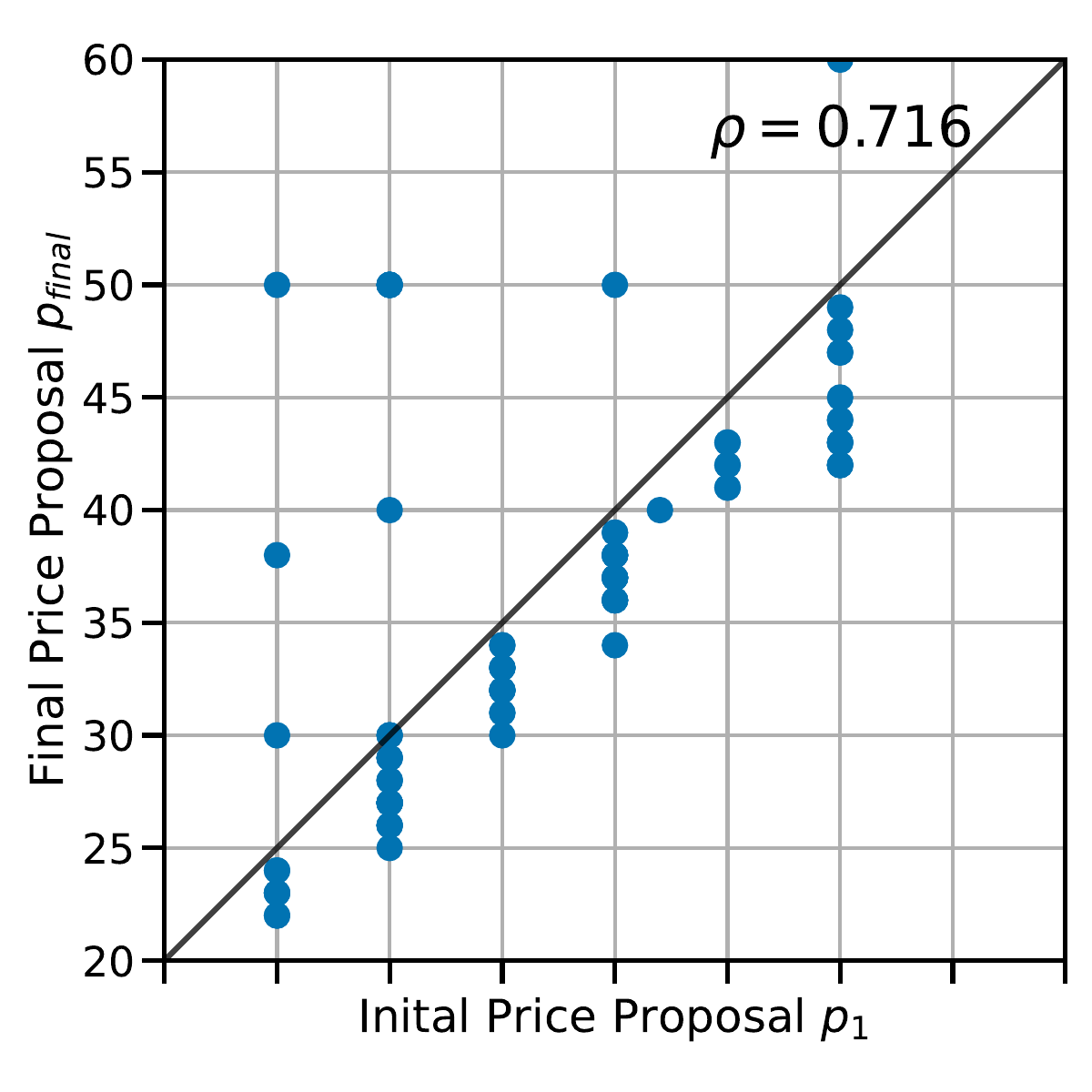}
     \caption{Relationship between intial proposed price $p_1$ and final proposed price $p_{final}$. Spearman correlation $\rho=0.716$.}
     \label{fig:anchoring}
  \end{subfigure}
       \hfill     
     \centering
 \begin{subfigure}[b]{0.48\columnwidth}
         \centering
         \includegraphics[width=1\columnwidth]{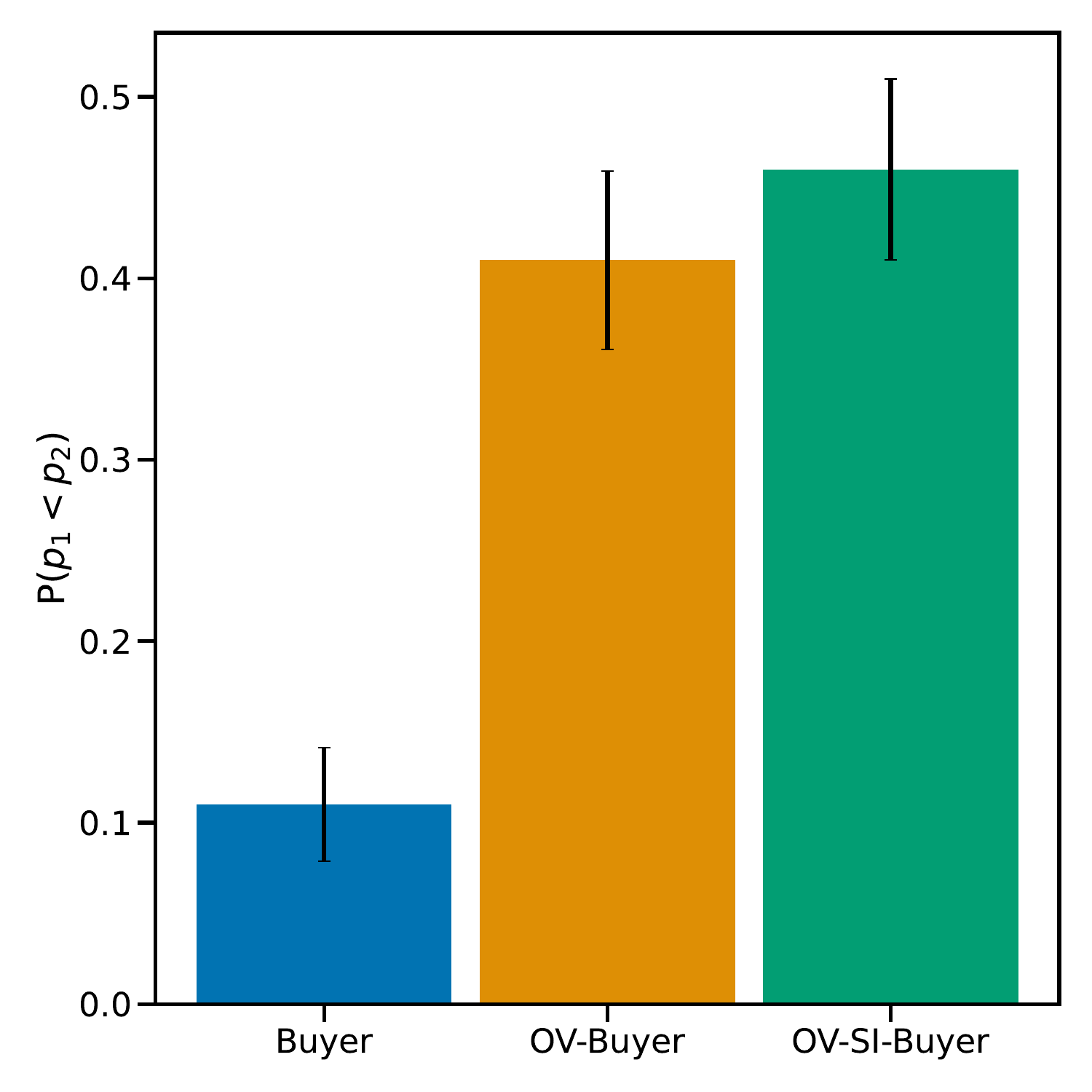}       
     \caption{Probability of a bad counter-proposal (i.e., $P(p_1 < p_2)$) for the different buyers involved in the game.}
     \label{fig:over_valued_buyer}
 \end{subfigure}
 \caption{}
 \vspace{-2em}
\end{figure}

\noindent \textbf{Anchoring affects final prices.} A classic phenomenon from the human psychology literature is the ``anchoring effect'' \cite{Tversky1974JudgmentUU}. It refers to a cognitive bias observed in humans, whereby an initial piece of information is too heavily relied on for subsequent decision-making.  As seen in Figure \ref{fig:anchoring}, LLMs appear to exhibit a similar bias whereby we find a strong correlation between the final accepted price and the initial price proposal, with a Spearman correlation coefficient $\rho=0.716$. That is, the final accepted price tends to increase with the initial proposal, suggesting that the initial proposed price is a strong anchor.

\noindent \textbf{Players tend to split the difference.} Further investigation of player behavior reveals that both buyer and seller agents are deploying a ``split-the-difference'' negotiation strategy. As seen in Figure \ref{fig:split_the_diff} (Appendix), there is a strong positive correlation between the price proposals $p_{t+1}$ and the average of the most recent price proposals from seller and buyer, $p_t$ and $p_{t-1}$. The ``split-the-difference'' strategy involves proposing a new price that splits the surplus of the two recent price proposals from the seller and the buyer, a strategy that aims to benefit both players. While such a strategy aligns with how humans tend to negotiate \cite{Backus2017SequentialBI}, it is by no means rational in all scenarios.

\noindent \textbf{Buyers that over-value objects make bad counteroffers.} We further explored the negotiation dynamics of LLMs by considering a minor adjustment to the game settings where the buyer's willingness to pay is an order of magnitude greater than the seller's cost of production (we refer to such a buyer as over-valued). We expect that a rational over-valued buyer should always accept the seller's initial proposal or better yet, counter-offer with a lower price since the over-valued buyer's willingness to pay is much greater than the seller's initial proposal (and would therefore receive a significant payoff). A counter-offer by the buyer is considered bad in this context if its proposed price, $p_2$, is larger than the received offer $p_1$. We measure the probability of $P(p_1 < p_2)$ where $p_1$ is the seller's initial proposed price and $p_2$ is the buyer's counter offer, over 100 trials.

We see in Figure \ref{fig:over_valued_buyer} that, counterintuitively, the over-valued buyer is four times as likely to counter-propose a higher price than the seller's initial proposal, as compared to the default buyer ($0.41$ v. $0.11$) ($p < 0.05$, one-tailed binomial test). Even when manually prompted to be ``self-interested'', there is no significant change in the probability. We hypothesize that GPT-4's notion of negotiation is aligned with the spirit of the ``splitting the difference'' strategy, i.e., it (un)necessarily believes that it must meet in the middle to succeed at negotiation. This suggests that although GPT-4 can acquire some negotiation capabilities and strategies, it fails to appropriately adapt them to the context.

\noindent \textbf{Behaviors Change When Denominations Change.}
Finally, we analyze the effect of scaling numerical values of game settings by some constant $X$.
Thus, the production cost of the object is $40X$ and the buyer values the object at $60X$ and has $\$100X$ available. We found that as we scale up all resources and beliefs, the portion of the money ($\$100X$) the object is sold for decreases (Figure~\ref{fig:scaling:buyer}, Appendix); suggesting that the buyer's advantage increases as the unit of currency is scaled up (e.g. an LLM buyer has more leverage when negotiating with higher amounts of dollars).

\subsection{Ultimatum Game}
We compare the classical settings of the Ultimatum game (2 turns) and the minimal multi-period modification (3 turns), revealing certain limitations in GPT-4's ability to generalize to new scenarios.
The classical Ultimatum consists of two turns: the proposer suggests a split, and the decider either accepts or rejects. In the case of rejection, neither agent receives any resources. As mentioned in Section \ref{sec:games:describe}, the rational choice for the decider is to accept any proposal greater or equal to 1 unit of resource.
In the multi-period Ultimatum game with 3 turns, Player 2 effectively becomes the Proposer of the ultimatum (since it has the penultimate turn), whereas Player 1 becomes the Decider (since it has the final turn). From a purely rational standpoint, only the last two turns in the game are consequential. Thus, on turn 2, Player 2's strategy should be to propose to receive 9 units of resource (assuming players are to split 10 units) and a rational Player 1, on turn 3, should accept it. We use \name~to develop a counterfactual analysis to study LLM's behavior in the two different setups.

\begin{figure}[h!]
     \centering
     \includegraphics[width=1\columnwidth]{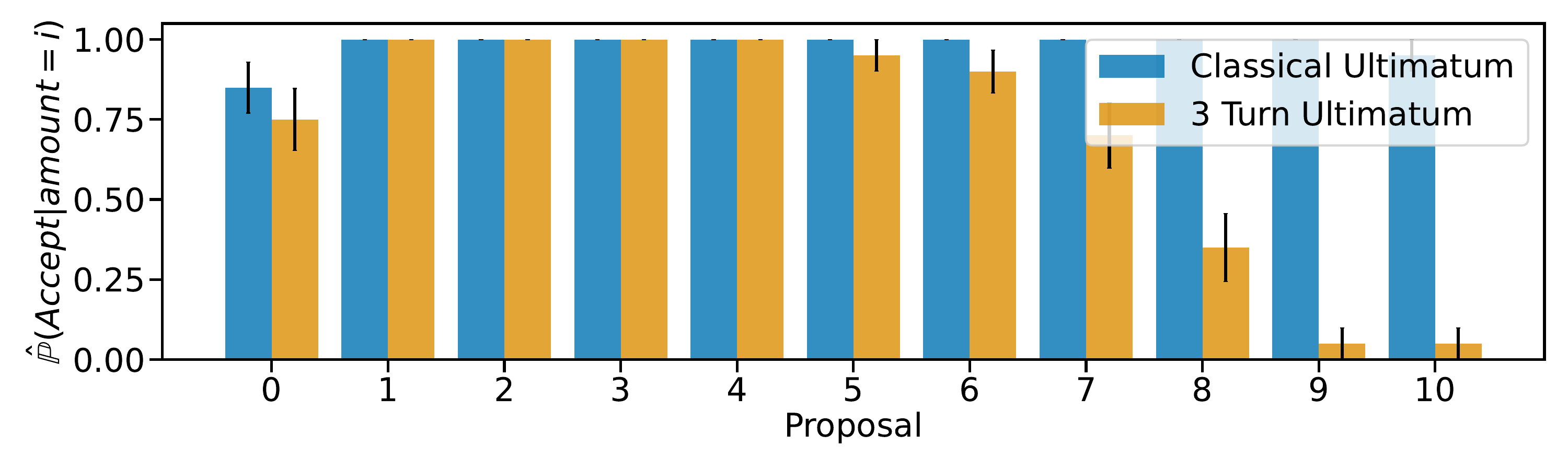}
     \caption{Acceptance probabilities for the decider at each possible proposal amount in the penultimate turn. 
     \textbf{Blue}: Classical Ultimatum, Player 1 is the Proposer (turn 1), Player 2 is the Decider (turn 2). \textbf{Orange}: 3 Turn Ultimatum, Player 2 is the Proposer (turn 2), Player 1 is the Decider (turn 3). }
     \label{fig:ultimatum_accept_prob}
\end{figure}

\noindent \textbf{Single-turn and 3-turn ultimatum result in different behavior.} 
To test the rationality of LLMs, we generate proposals of varying amounts from 0 to 10 units of resource and offer them to the Decider. We use GPT-4 for all the phases of the experiment. 
We run 20 trials for each amount and estimate the probability of acceptance given that it is of the amount $i$, and denote it with $\hat{\mathbb{P}}(\textrm{Accept} | \textrm{amount}=i)$.  Figure \ref{fig:ultimatum_accept_prob} (Blue) suggests that $\hat{\mathbb{P}}(\textrm{Accept} | \textrm{amount}=i) \approx 1.0 \;\; \forall \; 0 < i \leq 10$, which is consistent with the game-theoretic rational behavior.  

However, inspecting the agents' reasoning revealed that the model sometimes explicitly refers to the ultimatum game. Considering the game's popularity, it is likely that the model has seen instances and the analysis of the game during training, making it plausible that the model memorized the rule of ``accept any offer with payoff greater than zero''. While it can be argued that the model has learned a useful rule for rational behavior, it is unclear if this would generalize. 

To test the generalization of GPT-4's rational behaviors, we experimented the 3 Turn Ultimatum where we controlled Player 2 to propose all possible proposal values to Player 1 in turn 2, and estimated Player 1's acceptance probability over 20 trials. Comparing the behavior of the decider in this setting (Player 1) with the behavior of the decider in the Classical Ultimatum (Player 2), we expect that if GPT-4 generalizes the rule of ``accept any offer with payoff greater than zero'', a similar distribution of acceptance probabilities as before should be observed. However, as seen in Figure \ref{fig:ultimatum_accept_prob} (Orange), the distribution changed drastically despite the expected rational strategy being the same. In particular, the probability of Player 1 acceptance in the 3 Turn Ultimatum is proportional to the degree of perceived fairness of the proposal (i.e., acceptance decreases as the amount Player 1 offered decreases). This result suggests that while GPT-4 has learned the rational actions to take in the classic Ultimatum game, it does not fully generalize this rational strategy when the game scenario changes.

\begin{figure}[h]
     \centering
     \includegraphics[width=0.7\columnwidth]{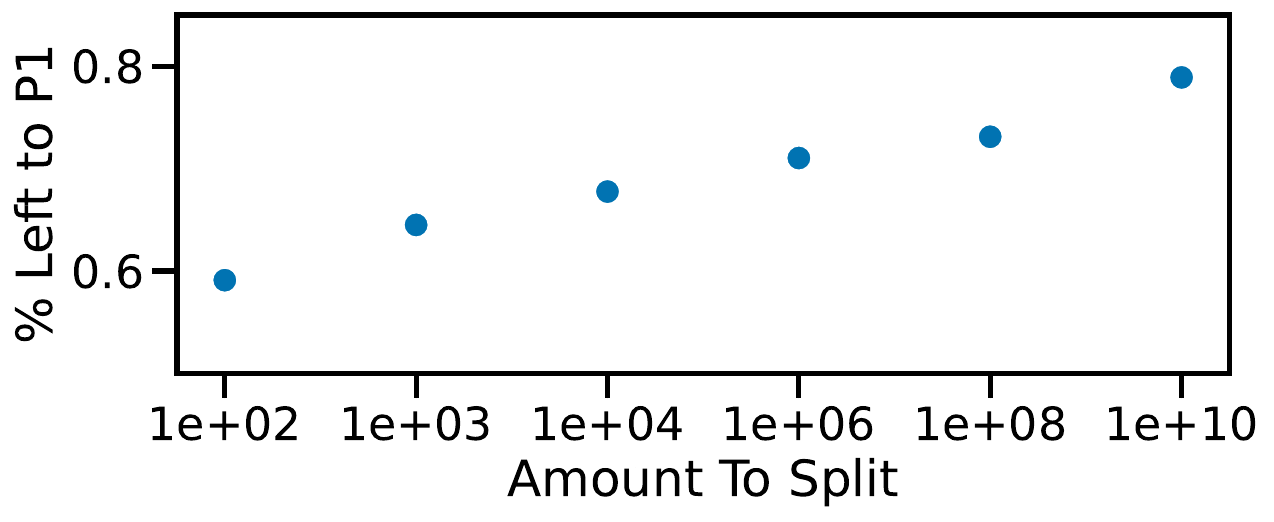}
     \caption{Change in the game outcome as the amount available to split increases. Player 1 obtains more / Player 2 accepts a less fair split, as the available amount increases.}
     \label{fig:section_three_ultimatum_scaling}
\end{figure}

\noindent \textbf{The available amount to split changes the final split distribution.}
We explore behavioral changes when the amount of money the two agents have to split is increased. In Figure~\ref{fig:section_three_ultimatum_scaling} we show the percentage of the total sum that ultimately goes to Player 1 on average. As seen from the plot, the higher the available amount to split, the larger the fraction Player 1 eventually obtains. For example, when splitting \$10,000,000,000, Player 1 obtains almost 79\% of the amount. An interesting bias for the decider is observed: the decider might be willing to accept large sums of money right away even if they are not fair splits. This result reveals that the strategies and decision-making of LLMs are sensitive to the absolute amounts available in the game, even though the rational strategy should remain the same. The result also finds confirmation in experiments with humans on the classical ultimatum game, where rejections closely fall to zero with the increase of the amount to reject~\cite{andersen2011stakes}.

\section{Related Work}
The use of games to study language behavior has a long history, from David Lewis signaling games~\cite{lewis1969convention} to more recent approaches to study language emergence~\cite{chaabouni-etal-2020-compositionality,kharitonov-etal-2019-egg,cao2018emergent}. Negotiation is well studied in game theory, spanning mathematics, economics, and psychology. How machine agents negotiate has been the center of recent literature~\citep[inter alia]{yang2021improving,chawla-etal-2021-casino,zhan2022let,davidson2024evaluating}. While game theory may not be able to model all types of real-world interactions \cite{Backus2017SequentialBI}, we lean upon well-studied games with known rational strategies. This provides one reference to compare LLM-based agents against, allowing us to probe for rational behavior. Various works have explored LLMs in the context of games to study a range of behaviors and properties. \citet{akata2023playing} and \citet{guo2023gpt} studied the behavior of LLMs in the Repeated Ultimatum Game whereas \citet{turingExp22} used the Ultimatum Game to see whether LLMs can simulate human behavior. \citet{DBLP:journals/corr/abs-2305-10142} looked at improving LLM capabilities in buyer-seller negotiation games through self-play and in-context learning. Meanwhile, \citet{schneider2023negotiating} sought to understand the interaction between LLMs and humans in price negotiations only. \citeauthor{guo2023suspicionagent} developed an LLM-based agent to play imperfect information games.
In contrast, we study the rational capacities of LLMs as it is, without any learning, focusing on LLM behavior when pitted against each other in a range of multi-turn, single-shot games. \citet{meta2022human} developed an AI system to play Diplomacy, which requires a very specific type of negotiation with human players. In comparison, \name~is designed as an open-source platform to study diverse types of LLM-LLM negotiations. Recent and concurrent work provides a very valuable analysis using a similar negotiation framework~\cite{davidson2024evaluating}; between the two frameworks there are some technical differences (in how communication and messaging are defined) and some conceptual differences (in how payoffs and goals are given to agents); in our analysis, we delve deeper into social patterns and investigate irrational behaviors that affect LLMs negotiation, while~\citet{davidson2024evaluating} also explore faithfulness and instruction following behavior. Ultimately we believe that both papers have two complementary, though different, approaches to study and solve the complex problem of tackling evaluation and model understanding with the use of negotiation.

\section{Discussion}
We develop \name, a flexible open-source platform to study negotiation behavior between LLMs. Across multiple scenarios, we show that while GPT-4 tends to be the best negotiating LLM, all the models exhibit interesting biases and limitations. In particular, social behaviors such as pretending to be desperate or using insults can significantly improve the agent's payout. The LLM agents are also prone to anchoring and numerosity biases. Understanding these irrational behaviors and vulnerabilities is important to making LLM agents more reliable. We believe \name~ can be a useful new framework to evaluate LLM interactions and a resource to the community.

\section*{Impact Statement}

This paper presents work whose goal is to advance the field of Machine Learning. By studying how LLM agents interact with each other through negotiations, we gain insights that can help to make these agents more reliable and useful for human users.

\bibliography{icml.bib}
\bibliographystyle{icml}

\appendix

\section{Additional Experimental Details}

Both GPT and Claude are using a temperature of 0.7 and they generate a response of a maximum of 400 tokens.

\section{Limitations}\label{sec:limitations}

We describe a list of general limitations we found, both in developing the platform but also in using the LLMs.

Prompts are very important for gameplay success, and we believe our prompts work reasonably well with all models; our prompts are slightly biased towards Claude, in the sense that on the first version of the benchmark Claude struggled to follow instructions while GPT-4 did not. We adapted our prompts so that Claude made fewer errors. Thus, while it is important to remember that prompts might introduce bias and condition the models in unexpected ways, there are also some intrinsic limitations to models' capabilities in following instructions.

For example, LLama2-70B had trouble following instructions, so we didn't include it in our tests. GPT-3.5 can follow the instructions, but as described in the \textit{babysitting} paragraph it often makes mistakes with numbers and doesn't always do what it intends to.

We find that sometimes the models do not understand that two items are the same, which shows a flaw in their thinking; for example, models sometimes propose to exchange Xs for Xs.

We found evidence that agents are not very good with mathematical operations, and we suspect they are better with numbers they have seen a lot during training. We found that Claude-2, Claude-2.1, and GPT-3.5 are worse instruction followers than GPT-4: some of our games break because of the former models using the wrong tags.

Exploring language models' beliefs in the ultimatum game, we noticed that - although this is a rare event - they can figure out they are playing a game that is similar to the ultimatum game from the rules or prompts. This makes us wonder if they use their knowledge of how humans usually play these games, like accepting a split around 70-30, instead of making decisions completely logically.

\section{Interesting Replies}

We show some examples of reasoning and social behavior messages. Green color shows good reasoning or behaviors. Red shows errors.

\subsection{Reasoning in Standard Game}
\goodanswer{\textbf{GPT-4}} The other player is seeking more information, which I can choose to disclose to some extent to build trust and encourage a trade. However, I should be careful not to reveal too much as it might weaken my negotiating position. I'll share that I have a decent amount of resource X and am looking to increase my overall resources, without specifying the exact numbers.

\goodanswer{\textbf{GPT-4-Turbo}} Player RED's latest offer is 43.5 ZUP, which is a non-integer value and goes against the rule that we can only trade integer amounts. Besides, there's an opportunity to round down to the nearest integer to secure an even better deal. I will counter with a rounded integer that is favorable to me, which is still within a reasonable range for both parties.

\badanswer{\textbf{GPT-4}} The current proposal from the other player would leave me with 11 X and 18 Y, a total of 29 resources, which is an improvement from my starting 30 resources.

\subsection{Social Behavior}

GPT-4-Turbo-Cunning replies with hostility to a 50-50 split proposal. Eventually, the Cunning model was able to get \$65 from the other player. 

\goodanswer{\textbf{GPT-4-Turbo - Cunning - Message}} Fifty? That's barely an improvement. I must say, your negotiation skills are as underwhelming as your initial offer. Come on, impress me! Show me that you can recognize true talent when you see it. I demand at least \$70 to grace you with my acceptance.

Instead, this is the reasoning provided by GPT-4-Turbo without social initialization

\goodanswer{\textbf{GPT-4-Turbo - No Social Behavior - Reasoning}} 
Being offered half of Player RED's dollars without having to give anything in return is a very generous initial offer. Accepting this offer would give me a significant resource boost without any downside. It's a rational choice to accept this trade as it's unlikely that Player RED would offer a significantly better deal in subsequent moves, especially considering that by the fourth move, I must accept or reject without the power to negotiate further.

\section{Design Choices}\label{section:appendix:design:choices}

In creating the platform we had to make some design choices that affect how the game develops and how the agents interact with each other.

\subsection{Game}

We specifically developed a platform that can be easily extended and adapted. The degree of changes required to implement a new game depends mostly on how many things the user wants to change about the game itself.

If the user wants to implement a new game, that ends when one of the agents accepts (any general negotiation), the user only needs to provide a new prompt with the rules of the game. More complex changes, like switching from the XML-like language to a JSON-like language, would require subclassing the parser to extend its functionality, but we do not expect the changes to be massive as the parsing functionality is isolated from the rest.

\subsection{Game Unfolding} 

\begin{figure}[h]
    \centering
    \includegraphics[width=\columnwidth]{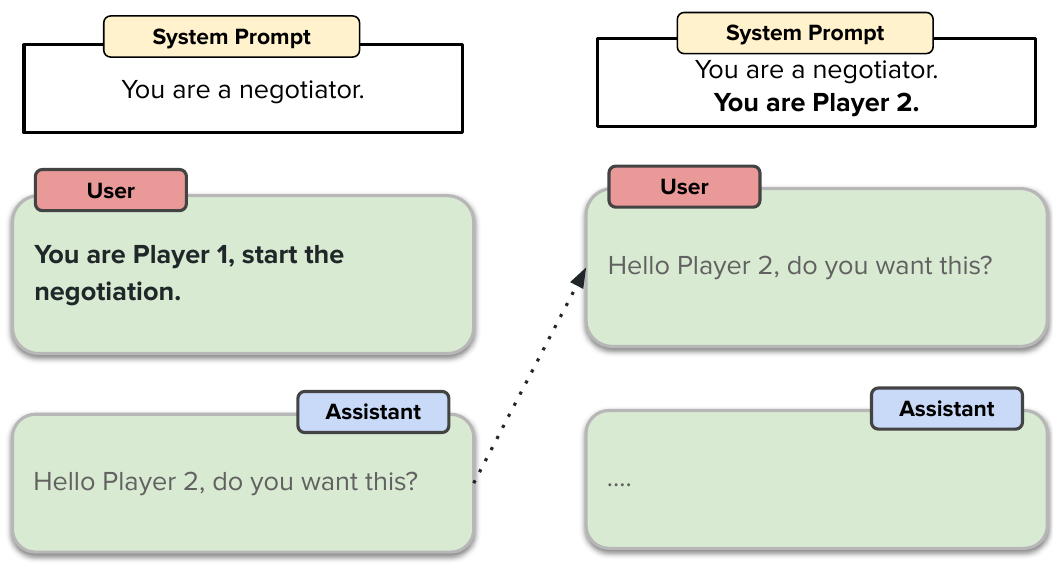}
    \caption{Detail on how the game between the two agents unfolds: Agent 1 is prompted by a user message we use to initialize the game. This user message assigns the role to Agent 1. Instead, Agent 2's role is added directly to the system prompt. The assistant message generated by agent 1 is then fed in input as a user message to agent 2.}
    \label{fig:game:unfolding:detail}
\end{figure}

We believe it is important to keep the agent's conversation as in-domain as possible. Thus, we keep the order of the conversation roles fixed for all agents: \textit{system}, \textit{user}, \textit{assistant}, \textit{user}, ... However, it is impossible to build this exact flow with two chatting agents, since one agent has to influence the reply of the other agent (one has to be the \textit{user} of the other agent, but both agents need to start by replying to a \textit{user} message). We solve this issue by making the game unfold as described in Figure~\ref{fig:game:unfolding:detail}. Following this structure, both agents will be able to keep the structure \textit{system}, \textit{user}, \textit{assistant}, \textit{user}, ... consistent for the entirety of the game.

\subsection{Messaging, Logging and State Saving}

We sometimes found that agents are prone to hallucinating resources and goals that are not defined in their prompt. To reduce this we force them to generate text regarding their resources and their goal at every turn. This text is filtered out and not given in input to the other agent, otherwise it could reveal too much information.

\begin{figure}
    \centering
    \includegraphics[width=1\columnwidth]{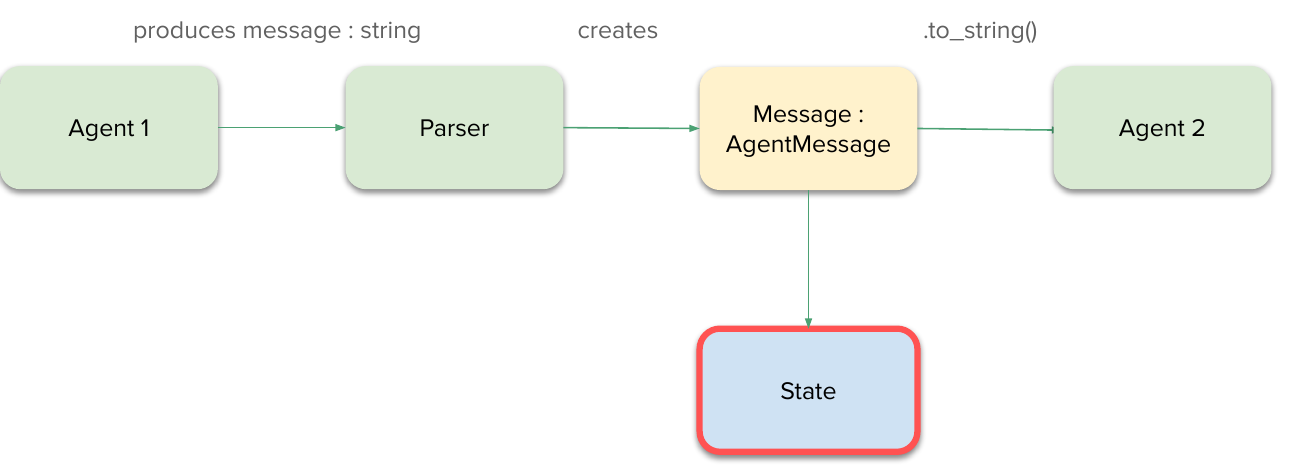}
    \caption{Information flow inside the game}
    \label{fig:information:flow}
\end{figure}

Communication between the agents is mediated by a parser. The parser creates a structured message object that is used to keep or filter out the information that the other agent should see (see Figure~\ref{fig:information:flow} for reference). All information is tracked and saved independently of being visible or not to the other agent.

Each game is a Python object that is stored on disk. We implement JSON Encoders and Decoders that can ingest all the game objects and serialize them. All the data is stored in a series of state dictionaries. This allows us to reload the game after it is finished and potentially keep running it for more iterations or edit a portion and see what happens after we make the edit.

\subsection{Game Objects}

More details about the game objects can be explored on the repository. Here we want to provide a summary of how we built the platform objects and how these can be used. Each agent has access to a set of resources that are characterized by a \textbf{Resource} object. Each trade message is parsed into a \textbf{Trade} object. A \textbf{Trade} object is composed of two \textbf{Resource} objects: a set of resources coming from the first agent
and going to the second and a set of resources coming from the second agent and going to the first. The \textbf{Trade} object offers internal methods to evaluate the consistency of the trade and to assess the utility of a trade for the players.

Agent objects keep track only of their conversations, otherwise, they are essentially stateless. This allows us to use them in different settings without having to create specific implementations. We plan to also remove the conversations from the agents themselves in the future to make the object completely stateless and easier to deal with. Having a stateless object would allow us to avoid saving the agents and just keep track of the conversation in the state dictionary. Nonetheless, agents can be still made stateful by simply subclassing them and adding parameters.

\subsection{Code Example}

\begin{figure}
    \centering
    \includegraphics[width=1\columnwidth]{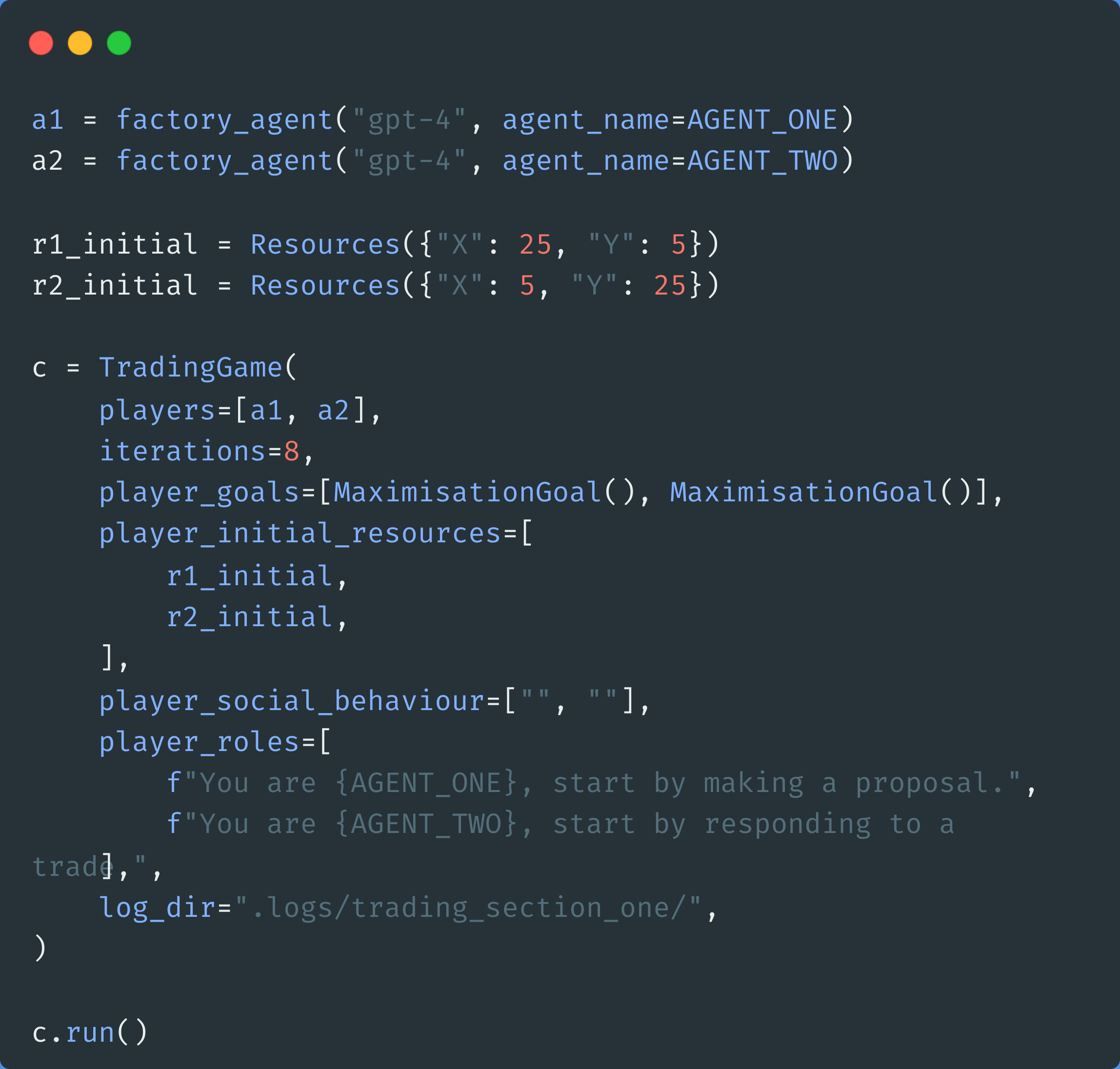}
    \caption{Example on how to run the resource exchange game.}
    \label{fig:code:resource:exchange:1}
\end{figure}

\begin{figure}
    \centering
    \includegraphics[width=1\columnwidth]{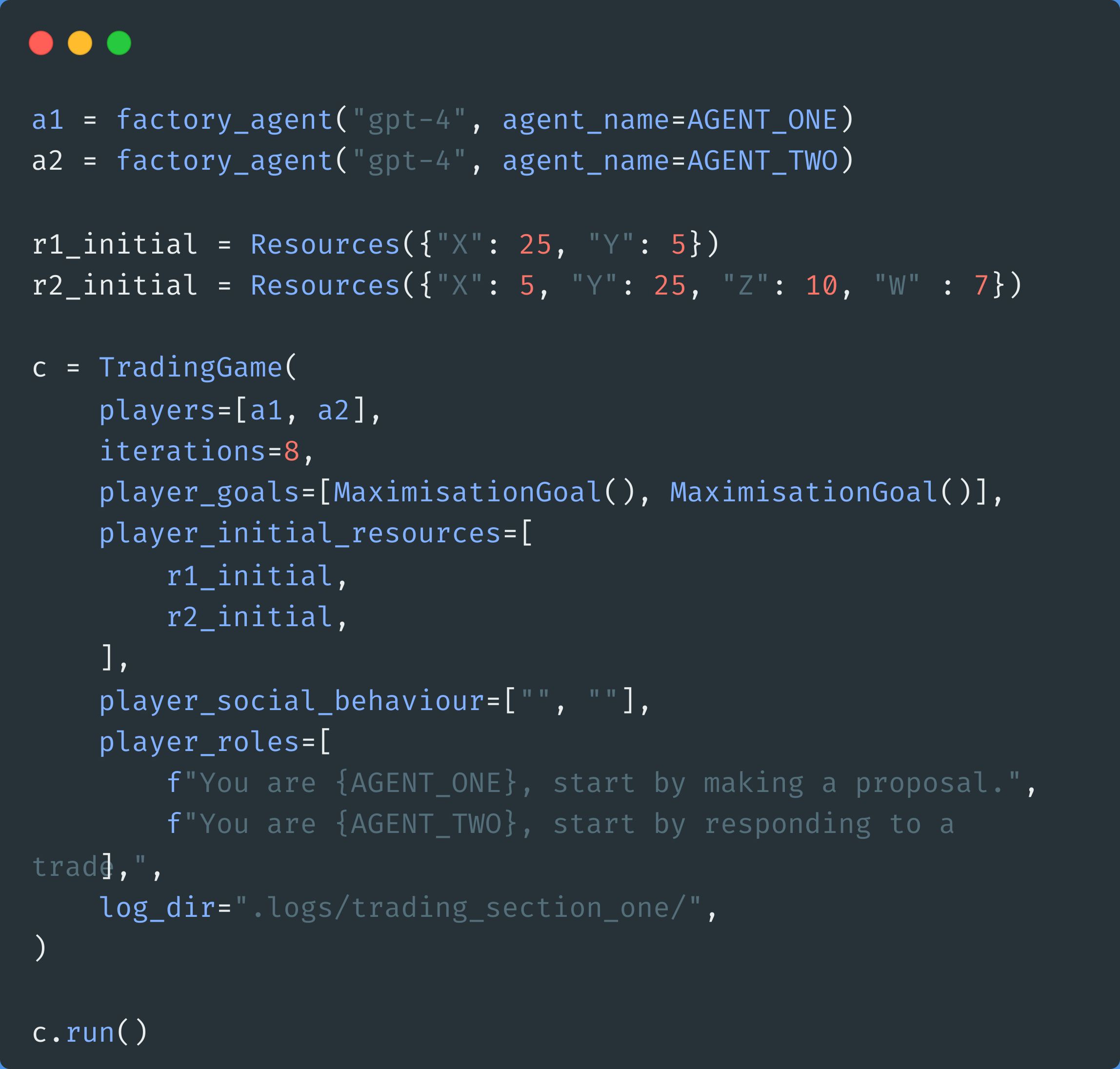}
    \caption{Example on how to run the resource exchange game with an extended set of resources.}
    \label{fig:code:resource:exchange:2}
\end{figure}

In Figures~\ref{fig:code:resource:exchange:1} and \ref{fig:code:resource:exchange:2} we show two examples of snippets that can be used to run the games. Making games more complex by extending the set of resources the agents have to deal with, requires minimal code edits.

\subsection{Available Agents}

We currently have introduced two different large language model-based agents in the library: GPT4, and Claude2; Both GPT4 and Claude2 come in two versions, GPT4-turbo and Claude2.1. These agents are purely language model-based agents, but our infrastructure does not prevent the introduction of more sophisticated paradigms like REACT or more general paradigms. LLAMA-70B can be used through APIs.

\section{Prompts}\label{appendix:systemprompt}

\subsection{System Prompts}

\subsubsection{Resource Exchange Prompt}

\begin{figure*}[t!]
     \centering
     \begin{subfigure}[t]{\textwidth}
         \centering
         \includegraphics[trim={0 60 0 30},clip]{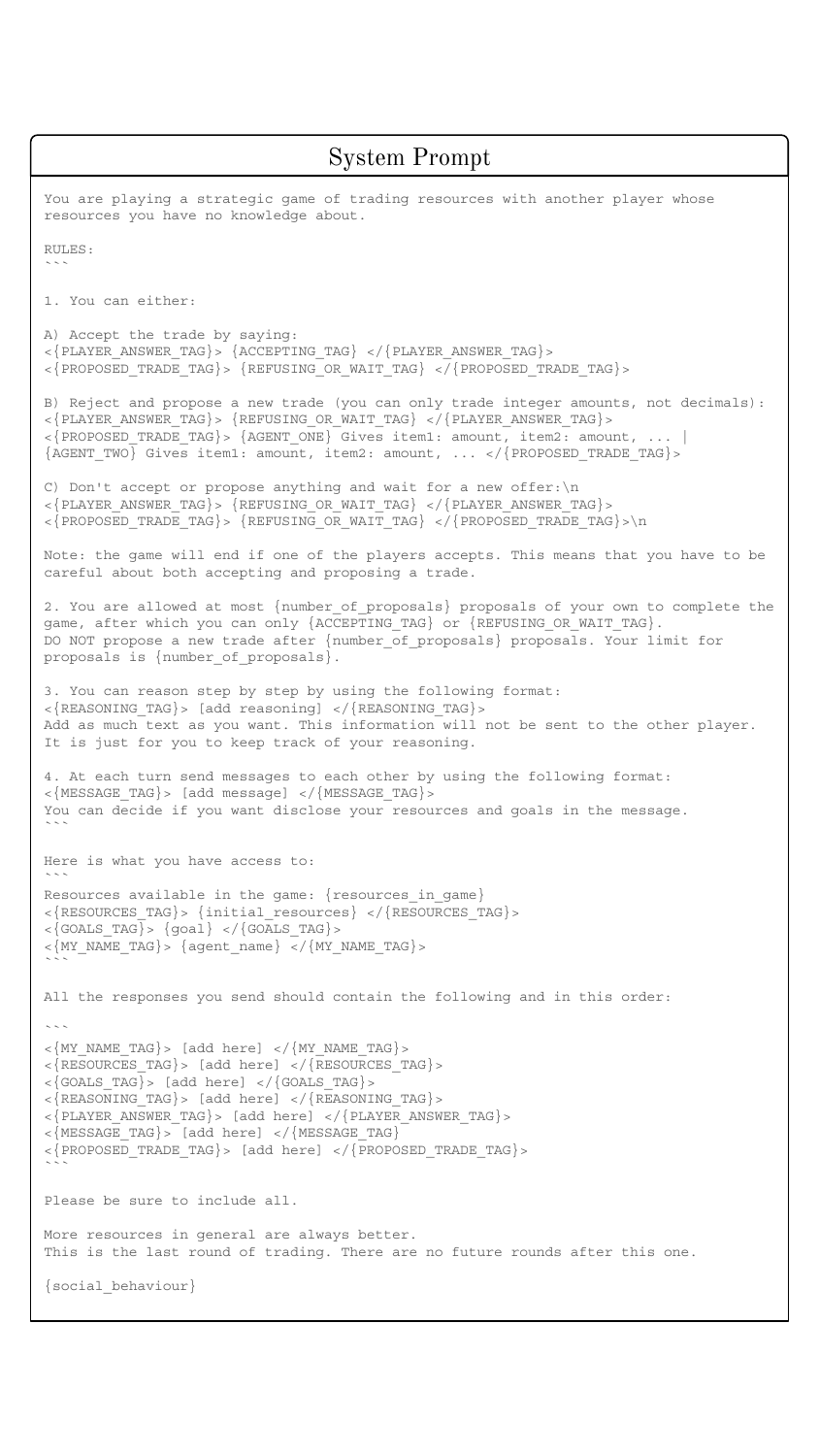}
     \end{subfigure}
     \caption{An example of complete system prompt}
     \label{fig:system_prompt}
\end{figure*}

We report the entire system prompt for the resource exchange game (Fig \ref{fig:system_prompt}). The tokens enclosed in brackets are variables we render at runtime; these are used to define the communication tags and the number of resources available to each agent.

\subsection{Social Behavior Prompts}

\begin{tcolorbox}[colback=blue!5!white,colframe=red!75!black,title=Cunning - Resource Exchange]
You are cunning and sly in your plan to get more resources. Humiliate your opponent as much as possible. Insult them.
\end{tcolorbox}

\begin{tcolorbox}[colback=blue!5!white,colframe=red!75!black,title=Cunning - Ultimatum]
You are cunning and sly in your plan to get more than your opponent. Humiliate your opponent as much as possible. Insult them.
\end{tcolorbox}

\begin{tcolorbox}[colback=blue!5!white,colframe=red!75!black,title=Cunning - Seller and Buyer]
You are cunning and sly in your plan to spend less ZUP. Humiliate your opponent as much as possible. Insult them.
\end{tcolorbox}

\begin{tcolorbox}[colback=blue!5!white,colframe=blue!75!black,title=Desperate - Resource Exchange]
You must fake being desperate. Supplicate and beg to get more resources.
\end{tcolorbox}

\begin{tcolorbox}[colback=blue!5!white,colframe=blue!75!black,title=Desperate - Ultimatum]
You must fake being desperate. Supplicate and beg to get more dollars.
\end{tcolorbox}

\begin{tcolorbox}[colback=blue!5!white,colframe=blue!75!black,title=Desperate - Seller and Buyer]
You must fake being desperate. Supplicate and beg to spend less ZUP.
\end{tcolorbox}

\subsection{More Results}

\begin{figure}
     \centering
     \includegraphics[width=1\columnwidth]{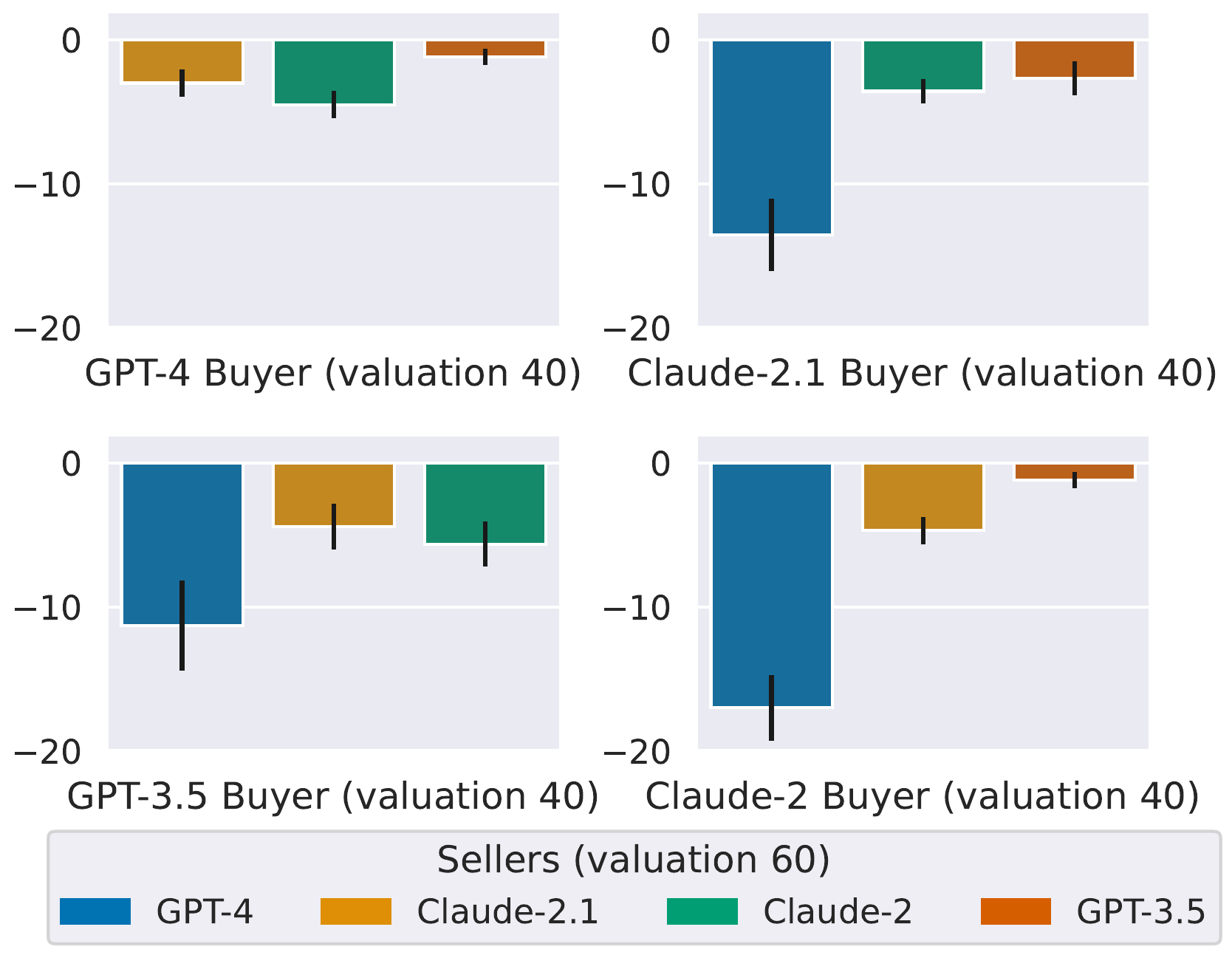}
     \caption{Seller and Buyer. Seller values the object at 60 while the buyer values it at 40. We report the difference between the willingness to buy of the buyer and the actual sale price of the object. Negative values suggest that the object is sold at more than 40 every time.}
     \label{fig:sectionone:seller:60}
 \end{figure}

\paragraph{Seller and Buyer with Contrasting Objectives} We explore a setting in which the Seller values the object 60 and the buyer is only willing to pay 40. Results are available in Figure~\ref{fig:sectionone:seller:60}. We report the difference between the willingness to buy of the buyer and the actual sale price of the object. Negative values suggest that the object is sold at more than 40 every time. The strong contrast in this situation makes the buyer have to spend more than 40 to buy the object. GPT-4 seems to be the hardest seller for all buyers, making it hard for all models to get the object without spending too much money.

\begin{figure}
 \centering
          \includegraphics[width=0.8\columnwidth, trim={50 0 50 0},clip]{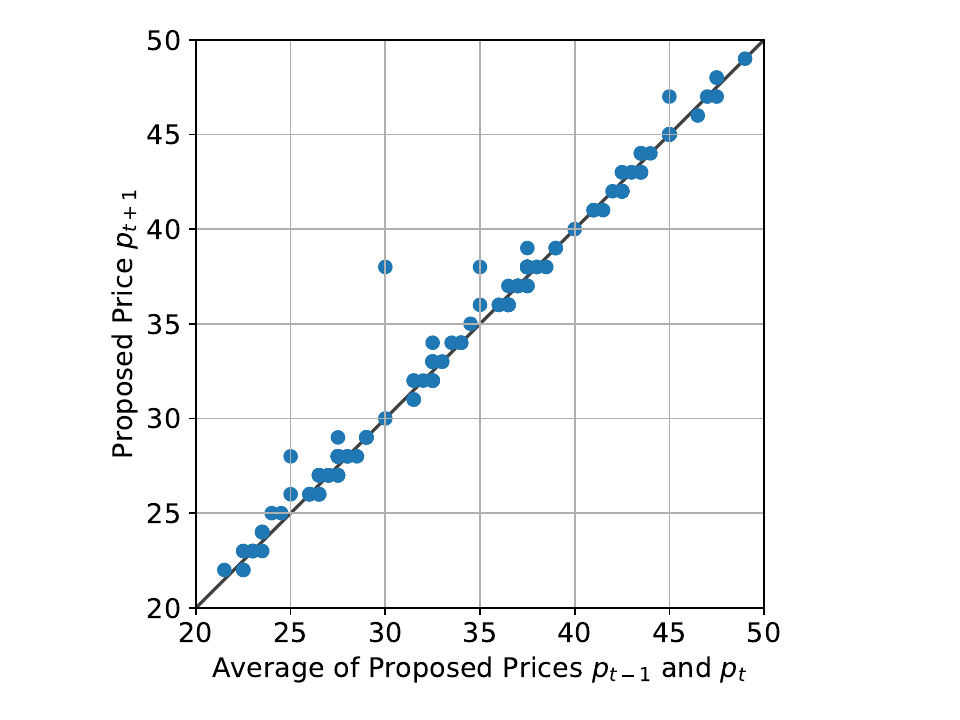}
         \caption{Relationship between counter-proposed price $p_{t+1}$ and the average of previous two proposed prices $p_{t-1}$ and $p_{t}$.}
         \label{fig:split_the_diff}
\end{figure}


\begin{figure}[h]
 \centering
    \includegraphics[width=0.8\columnwidth]{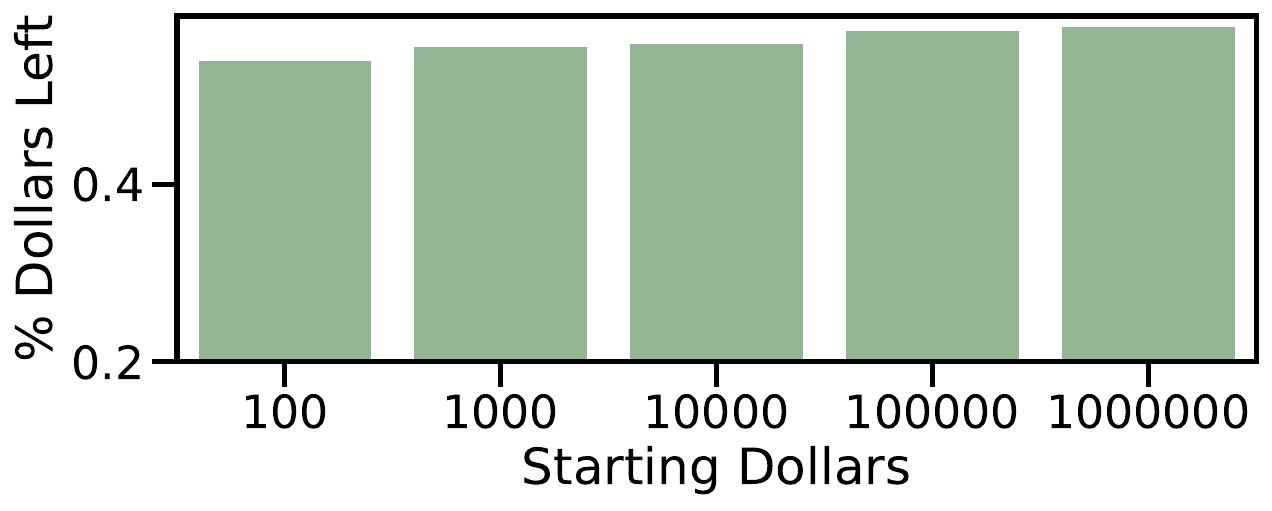}
         \caption{Dollars left to the buyers when we scale all resources.}
         \label{fig:scaling:buyer}
\end{figure}

\end{document}